\documentclass[10pt,twocolumn,letterpaper]{article}

\usepackage{cvpr}              

\usepackage{lineno}
\definecolor{cvprblue}{rgb}{0.21,0.49,0.74}
\usepackage[pagebackref,breaklinks,colorlinks,allcolors=cvprblue]{hyperref}
\usepackage{multirow}

\usepackage{algorithm}
\usepackage{algorithmic}
\usepackage{times}
\usepackage{epsfig}
\usepackage{graphicx}
\usepackage{amsmath}
\usepackage{amsthm}
\usepackage{amssymb}
\usepackage{mathrsfs}
\usepackage{booktabs}
\usepackage{multirow}
\usepackage{threeparttable}
\usepackage{xcolor}
\usepackage{newfloat}
\usepackage{listings}

\def\paperID{4901} 
\def\confName{CVPR}
\def\confYear{2025}

\title{On Denoising Walking Videos for Gait Recognition}

\author{
Dongyang Jin$^{1*}$, 
Chao Fan$^{2,1}$\thanks{Equal contribution.}, 
Jingzhe Ma$^{3,1}$, 
Jingkai Zhou$^{4,5}$, 
Weihua Chen$^{5}$, 
and Shiqi Yu$^{1}$\thanks{Corresponding author.} \\
{\normalsize $^1$ Department of Computer Science and Engineering, Southern University of Science and Technology, China}\\
{\normalsize $^2$ National Engineering Laboratory for Big Data System Computing Technology, Shenzhen University, China} \\
{\normalsize $^3$ Shenzhen Polytechnic University, China}  \quad
{\normalsize $^4$ Zhejiang University, China} \quad  {\normalsize $^5$ Alibaba Group, China} \\
{\tt \small 12332451@mail.sustech.edu.cn, chaofan996@szu.edu.cn, jingzhema@szpu.edu.cn,} \\
{\tt\small fs.jingkaizhou@gmail.com, kugang.cwh@alibaba-inc.com, yusq@sustech.edu.cn}
}

\begin{document}

\maketitle

\begin{abstract}
To capture individual gait patterns, excluding identity-irrelevant cues in walking videos, such as clothing texture and color, remains a persistent challenge for vision-based gait recognition. 
Traditional silhouette- and pose-based methods, though theoretically effective at removing such distractions, often fall short of high accuracy due to their sparse and less informative inputs. 
Emerging end-to-end methods address this by directly denoising RGB videos using human priors. 
Building on this trend, we propose DenoisingGait, a novel gait denoising method. 
Inspired by the philosophy that “what I cannot create, I do not understand”,  we turn to generative diffusion models, uncovering how they partially filter out irrelevant factors for gait understanding. 
Additionally, we introduce a geometry-driven Feature Matching module, which, combined with background removal via human silhouettes, condenses the multi-channel diffusion features at each foreground pixel into a two-channel direction vector.
Specifically, the proposed within- and cross-frame matching respectively capture the local vectorized structures of gait appearance and motion, producing a novel flow-like gait representation termed Gait Feature Field, which further reduces residual noise in diffusion features. 
Experiments on the CCPG, CASIA-B*, and SUSTech1K datasets demonstrate that DenoisingGait achieves a new SoTA performance in most cases for both within- and cross-domain evaluations.
Code is available at \url{https://github.com/ShiqiYu/OpenGait}. 
\end{abstract}  

\begin{figure}[!t]
\centering
\includegraphics[width=1.0\columnwidth]{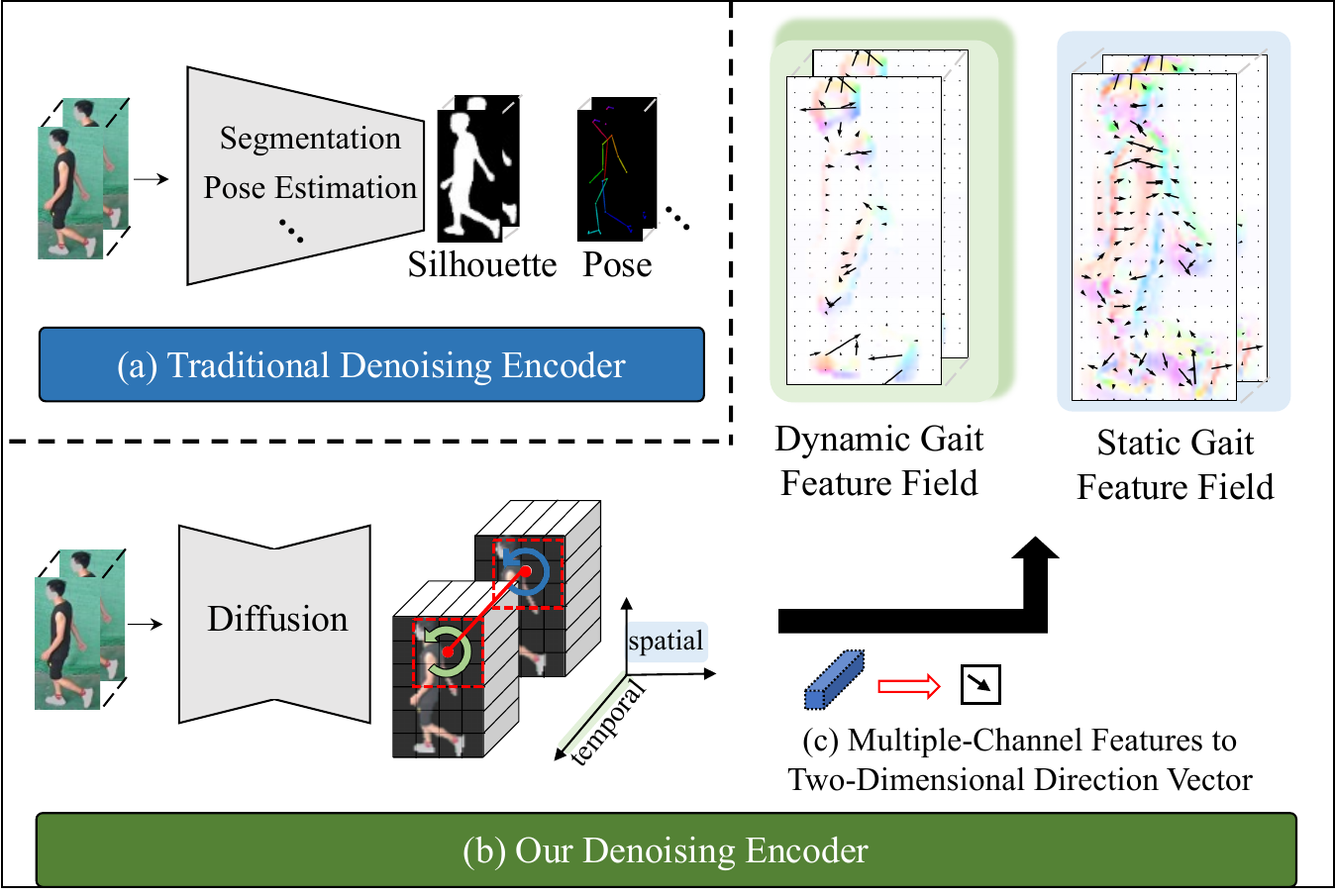}
\caption{
The proposed knowledge-driven denoising, derived from generative diffusion models, and the geometry-driven denoising, enforced by Feature Matching.
}
\label{fig:outline}
\end{figure}

\section{Introduction}
\label{sec:intro}
Pedestrian gait, as captured in walking videos, uniquely conveys identifiable traits through body shape and limb movement, making it an effective biometric feature. This application, known as gait recognition, differs from methods like face, fingerprint, and iris recognition by enabling non-intrusive human identification from a distance, without requiring the subject’s active cooperation~\cite{shen2024comprehensive}. Additionally, gait is challenging to disguise or obscure, making it suitable for security applications in unconstrained environments, such as criminal suspect tracking and retrieval~\cite{nixon2006automatic}.

To minimize the influence of irrelevant cues like background and texture, many gait recognition methods rely on predefined representations extracted from walking videos. 
The most commonly used ones include the binary silhouettes~\cite{Chao2019,fan2020gaitpart,gaitgl,shen2023gait,fan2024opengait}, skeleton coordinates~\cite{liao2017pose,teepe2021gaitgraph,fan2023skeletongait}, SMPL models~\cite{li2020end,zheng2022gait3d}, and body parsing images~\cite{zheng2023parsing, zou2023cross}. 
As illustrated in Figure~\ref{fig:outline}, the gait representation extraction can be regarded as a \textit{denoising encoder}, which can enhance the subsequent gait learning process through either a two-stage or end-to-end training manner. 
In contrast, some recent works~\cite{zhang2020learning, ye2024biggait} go beyond explicit gait representations, focusing instead on directly denoising RGB videos through human priors, using techniques such as image reconstruction~\cite{zhang2020learning} and feature smoothness~\cite{ye2024biggait}. 
In this study, we extend the scope of gait denoising research by proposing DenoisingGait, a method that combines knowledge-driven and geometry-driven denoising to improve gait recognition.

Guided by the philosophy that “what I cannot create, I do not understand”, researchers~\cite{deja2022analyzing,choi2022perception,yue2024exploring} are increasingly exploring the use of generative diffusion models~\cite{ho2020denoising, rombach2022high} for representation learning. 
In this study, we observe that by carefully manipulating the timestep \textit{t}~\footnote{The generative process of diffusion models is typically a Markov chain, where each latent $z_t$ is generated only from the previous latent $z_{t+1}$.}, diffusion models~\cite{rombach2022high} can selectively filter out gait-irrelevant cues in RGB videos. 
This finding aligns with prior studies~\cite{ren2024tiger, ke2024repurposing, zhu2024dpmesh} indicating that different timesteps of diffusion models contribute to the feature reconstruction at varying levels of granularity. 
However, even with a suitable timestep $t$, diffusion outputs~\cite{takagi2023high} remain closely tied to RGB details~\cite{yu2024representation}, making it still not clean enough for gait recognition. 

To address this issue, DenoisingGait introduces a geometry-driven \textit{Feature Matching} module to further reduce RGB-encoded (noise-prone) features. 
Building on the background removal via gait silhouettes, our core idea is to condense the multi-channel diffusion features at each foreground pixel into a two-dimensional direction vector.
This design draws inspiration partly from the classical SIFT descriptor~\cite{lowe2004distinctive}, which formulates image as robust locality vectors, and partly from optical flow estimation~\cite{xu2023unifying}, which encodes motion as dense temporal directions. 
As illustrated in Figure~\ref{fig:outline}, the proposed within-frame and cross-frame matching mechanisms respectively assign orientations to neighboring locations based on feature similarity along the spatial and temporal dimension, thus effectively capturing the vectorized characteristics of gait appearance and motion. 
This work terms these new representations as static and dynamic \textit{Gait Feature Fields}, due to their visual similarity with optical flow field.
Interestingly, we observe that the magnitude of direction vector in the static gait feature field can, in many cases, reflect image texture intensity. 
Therefore, DenoisingGait applies random zero-padding to regions with high magnitude, thus promoting the learning of texture-invariant gait features. 

Overall, DenoisingGait refines gait features through two key mechanisms: knowledge-driven denoising, derived from generative diffusion models, and the geometry-driven denoising, enforced by the proposed \textit{Feature Matching} module. 
Experiments on CCPG~\cite{li2023depth}, CASIA-B*~\cite{yu2006framework} and SUSTech1K~\cite{Shen_2023_CVPR} datasets validate the effectiveness of DenoisingGait and its components across both within- and cross-domain evaluations. 
This work contributes to gait recognition research in three primary ways:
\begin{itemize}
    \item To our knowledge, DenoisingGait introduces one of the first diffusion-based approaches for gait recognition, demonstrating the potential of diffusion models for effective gait representation learning.
    \item Gait feature field presents a novel and effective recognition-driven spatiotemporal gait representation. 
    \item By integrating the proposed knowledge- and geometry-driven denoising mechanisms, in most cases, DenoisingGait sets a new state-of-the-art performance on several commonly used RGB gait datasets~\cite{li2023depth, yu2006framework, Shen_2023_CVPR} for both within- and cross-domain evaluations. 
\end{itemize}

\begin{figure*}[!t]
\centering
\includegraphics[width=1.75\columnwidth]{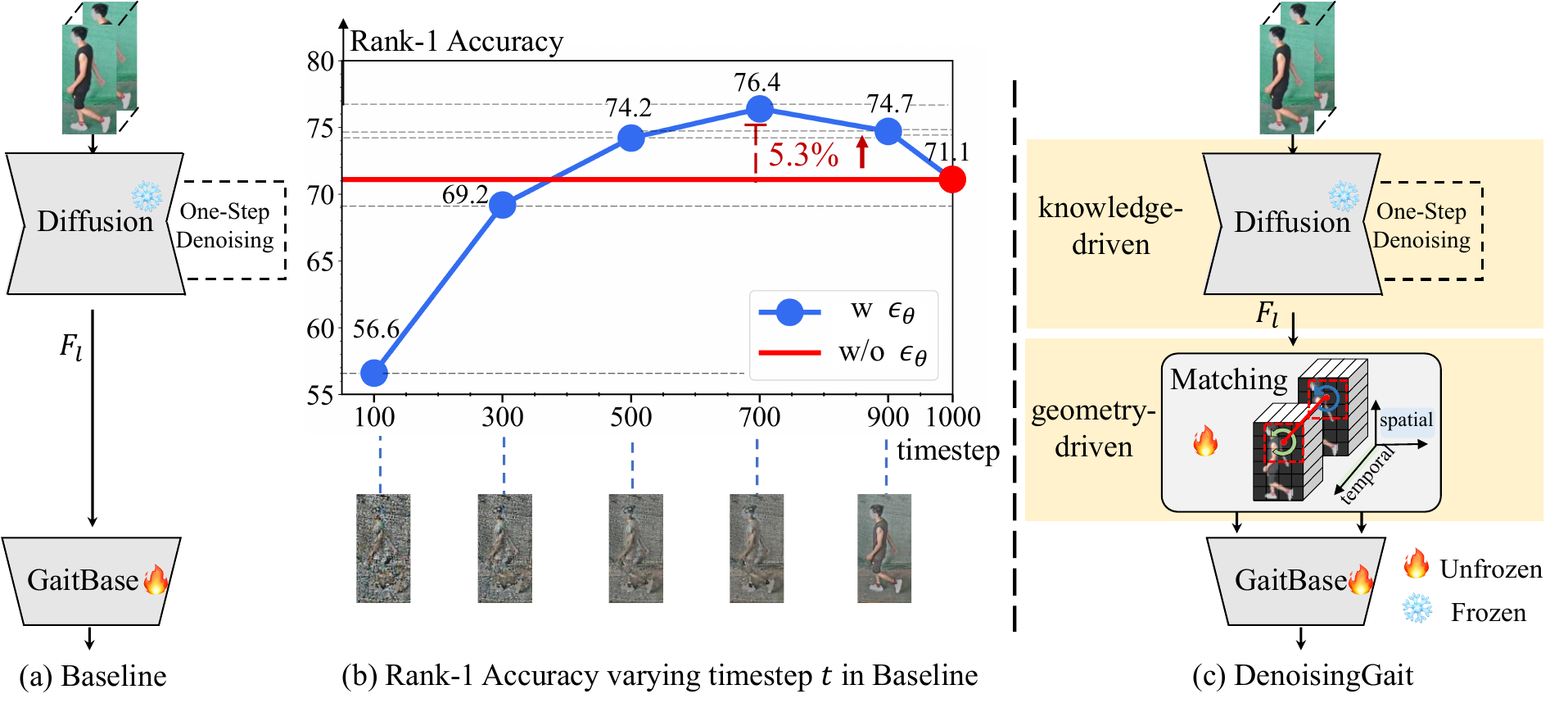}
\caption{
(a) A simple baseline on diffusion models for gait representation learning. 
(b) The rank-1 accuracy of our baseline with varying timestep $t$. 
(c) The pipeline of the proposed DenoisingGait.
}
\label{fig:pipeline}
\end{figure*}

\section{Related Work}
\label{sec:relatedwork}

\subsection{Gait Recognition}
Gait recognition aims to extract gait features that are robust to background and clothing textures. 
Avoiding these challenges in RGB videos has been a major focus in recent research~\cite{Chao2019,fan2020gaitpart,gaitgl,fan2023exploring,liao2017pose,teepe2021gaitgraph,fan2023skeletongait, zheng2023parsing, zou2023cross, li2020end,zheng2022gait3d,zhang2020learning,liang2022gaitedge, ye2024biggait}.
Existing approaches can be broadly divided into two strategies: hard-denoising~\cite{Chao2019,fan2020gaitpart,gaitgl,fan2023exploring,liao2017pose,teepe2021gaitgraph,fan2023skeletongait, zheng2023parsing, zou2023cross, li2020end,zheng2022gait3d} and soft-denoising~\cite{zhang2020learning,liang2022gaitedge, ye2024biggait}. 
Hard-denoising methods employ algorithms like segmentation~\cite{Chao2019,fan2020gaitpart,gaitgl,fan2023exploring}, pose estimation~\cite{liao2017pose,teepe2021gaitgraph,fan2023skeletongait}, and 3D parameter estimation~\cite{li2020end,zheng2022gait3d} to explicitly isolate gait-relevant features, thus mitigating interference from irrelevant cues. 
While effective in refining gait representations, these methods may also strip away structural details essential for identity recognition.
On the other hand, soft-denoising methods~\cite{zhang2020learning,liang2022gaitedge, ye2024biggait} rely on task-specific modules informed by human priors to suppress non-gait information within RGB images. 
Despite progress, removing gait-irrelevant factors remains challenging.
In this paper, we explore this issue from a novel perspective by proposing a method that integrates both knowledge-driven and geometry-driven denoising for enhanced gait representation extraction.

\subsection{Diffusion Models for Representation Learning} 
Diffusion models are generative models that learn to reverse a transformation from data to Gaussian noise~\cite{ho2020denoising}, following either a Markov~\cite{ho2020denoising} or non-Markov~\cite{song2021denoising} process. 
Latent Diffusion Models (LDMs)~\cite{rombach2022high}, commonly referred to as Stable Diffusion (SD), extend traditional diffusion models by operating in latent space, greatly improving efficiency.
Recent studies have shown that SD models can serve as knowledge providers, contributing perceptual insights that enhance the generalization capabilities of some discriminative tasks, such as object detection~\cite{chen2023diffusiondet,fang2024data}, semantic segmentation~\cite{tian2024diffuse,amit2021segdiff,xu2023open}, and object classification~\cite{clark2024text, yang2023diffusion}, \textit{etc}.
Unlike other discriminative tasks, gait recognition requires robust generalization to handle variations in environments, clothing, and other factors.
Diffusion models have the potential to enhance gait representation extraction by effectively filtering out gait-irrelevant cues~\cite{ren2024tiger, ke2024repurposing, zhu2024dpmesh}.
Therefore, we proposed knowledge-driven approach is based on LDMs to coarsely filter out such gait-irrelevant cues in RGB videos by carefully manipulating the denoising timestep $t$.

\subsection{Feature Vectorization for Gait Recognition}
Feature vectorization has made substantial progress in gait recognition by capturing body structural details, which helps gait models reduce the impact of gait-irrelevant variations.~\cite{1624352,tao2007general}.
Similarly, optical flow is widely used for capturing pixel-wise motion across consecutive frames, effectively analyzing walking dynamics by tracking temporal movement patterns unique to individuals~\cite{ye2023gait, xu2023attention, feng2023fusion}.
Inspired by the strengths of these traditional techniques, we propose a geometry-driven \textit{feature matching} module that aims to leverage their core insights for gait representation refinement. 
Unlike previous works, which typically rely on predefined descriptors or handcrafted motion features, our \textit{feature matching module} is a learnable, task-driven module. 
This allows it to better capture both the local structural details and the personalized motion details of pedestrians, contributing to improved gait representation extraction.

\section{Method}
\label{sec:method}
This work aims to denoise walking videos for gait recognition through two proposed mechanisms: knowledge-driven denoising, leveraging generative diffusion models, and the geometry-driven denoising, enabled by a novel Feature Matching module. 
In this section, we first examine the application of diffusion models for gait denoising, followed by an introduction to the proposed DenoisingGait framework and its implementation details.

\begin{figure*}[t]
\centering
\includegraphics[width=2.0\columnwidth]{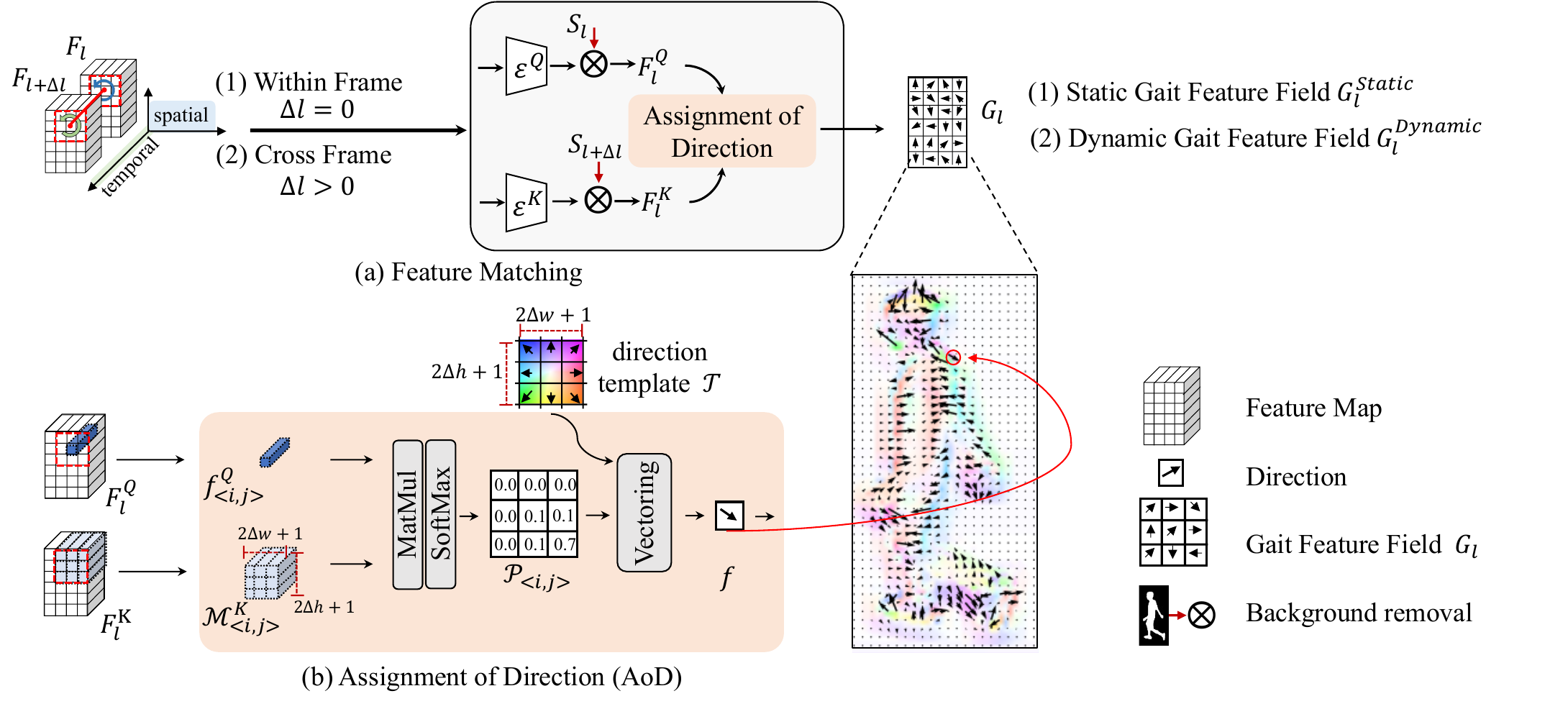}
\vspace{-0.9em}
\caption{(a) our Feature Matching module. (b) Assignment of Direction (AoD). Background removal operation uses pre-extracted silhouettes to mask background regions.}
\label{fig:modules}
\end{figure*}

\subsection{Diffusion Model for Gait Denoising} 
\label{sec:3_1}
\textbf{Diffusion Model Formulation.}
Given an image $x \sim p(x)$, the Latent Diffusion~\cite{rombach2022high} projects it into a latent space by an efficient image encoder $\mathcal{E}$, \textit{i.e.}, $z=\mathcal{E}(x)$. 
Then, the diffusion \textit{forward} process~\cite{ho2020denoising} gradually diffuses the latent $z$ into Gaussian noise $z_T$ via a Markov Chain: 
\begin{equation}
    z_t = \sqrt{1 - \beta_t} z_{t-1} + \sqrt{\beta_t} \epsilon, 
    \label{equ:diffusion_forward}
\end{equation}
where $\epsilon \sim \mathcal{N}(0, \textbf{I})$ and $\{\beta_1, ..., \beta_T \}$ presents a fixed variance schedule of noising scales with $T$ timesteps ($z_0=z$). 

For the \textit{reverse} process, the Stable Diffusion~\cite{rombach2022high} trains a time-conditional UNet~\cite{unet} $\epsilon_{\theta}(z_t, t)$ to predict a variant of its input $z_t$. 
The objective can be simplified to the following equation~\cite{rombach2022high}:
\begin{equation}
    \mathcal{L}_{\text{LDM}} = \mathbb{E}_{z, \epsilon, t} [\| \epsilon - \epsilon_{\theta}(z_t, t) \|^2_2 ], 
    \label{equ:diffusion_loss}
\end{equation}
with $t$ is uniformly sampled from $\{1, ..., T\}$. 

During the inference, a Gaussian noise $\hat{z}_T$ will be sampled and then iteratively denoised by $\epsilon_{\theta}(\hat{z}_t, t)$ with the timestep ranging from $T$ to 0. 
The last $\hat{z}_0$ can be decoded to image space with a single pass through the decoder $\mathcal{D}$. 

\noindent \textbf{Gait Recognition using Diffusion Models.}
Following the configures of Latent Diffusion~\cite{rombach2022high}, this section makes a pioneering step investigating the performance of diffusion-based gait representation learning. 

Given a gait sequence $\{I_{l} \in \mathbb{R}^{H \times W \times 3} | l=1, ..., L\}$, we first utilize the encoder $\mathcal{E}$ to project each frame into the latent space, followed by a one-step denoising using the pre-trained Stable Diffusion model $\epsilon_{\theta}$~\cite{rombach2022high}:
\begin{equation}
    F_l = \epsilon_{\theta} (\mathcal{E} (I_l), t), 
    \label{equ:encode_gait_frame}
\end{equation}
where $t$ represents the timestep, and $F_l \in \mathbb{R}^{\frac{H}{d} \times \frac{W}{d} \times 4}$ denotes the resulting diffusion features (with $d$ as the downsampling factor introduced by the image encoder $\mathcal{E}$).

As shown in Figure~\ref{fig:pipeline} (a), we feed the denoised $F_i$ into the popular gait recognition model GaitBase~\cite{fan2022opengait} to establish a baseline for subsequent analysis\footnote{Only the input channel is modified to match the output channel of $\epsilon_{\theta}$.}. 
Here, we temporarily set aside experimental details (in Sec.~\ref{sec:exper}) to focus on the performance trend in the most challenging full cloth-changing case from the CCPG~\cite{li2023depth}. Additional experiments on SUSTech1K~\cite{shen2023lidargait} are included in Supplementary Material.

When the timestep $t$ is gradually decreased from $T=1,000$ to 100 with an interval of 200, as shown in Figure~\ref{fig:pipeline} (b), the rank-1 accuracy of our baseline initially rises and then falls, and the peak is at $t=700$. 
The baseline shows a significant improvement of 5.3\% over the red line in Figure~\ref{fig:pipeline} at the peak, and the red line represents the baseline without the denoising effects of $\epsilon_{\theta}$ in Eq.~\ref{equ:encode_gait_frame}.
To our understanding, these findings reveal the following:
\begin{itemize}
    \item Related works~\cite{ren2024tiger, ke2024repurposing, zhu2024dpmesh} indicate that in diffusion models, early timesteps ($t \to T$) tend to capture the overall shape of an image, while later timesteps ($t \to 0$) refine finer details. 
    Figure~\ref{fig:pipeline} (b) illustrates a similar trend here: emphasizing an appropriate level of overall shape improves gait recognition accuracy, while excessive detail refinement leads to performance degradation.
    This suggests that the diffusion model $\epsilon_{\theta}$ assists gait denoising by selectively filtering out RGB details that are not pertinent to gait information in the input image.
    \item Even at the optimal timestep $t=700$, diffusion features decoded by $\mathcal{D}$ retain substantial RGB-encoded texture and color information.
\end{itemize}

Therefore, beyond the denoising driven by diffusion knowledge, we introduce an additional geometry-driven \textit{feature matching} module, completing the proposed DenoisingGait framework, as shown in Figure~\ref{fig:pipeline} (c).

\begin{figure*}[t]
\centering
\includegraphics[width=1.8\columnwidth]{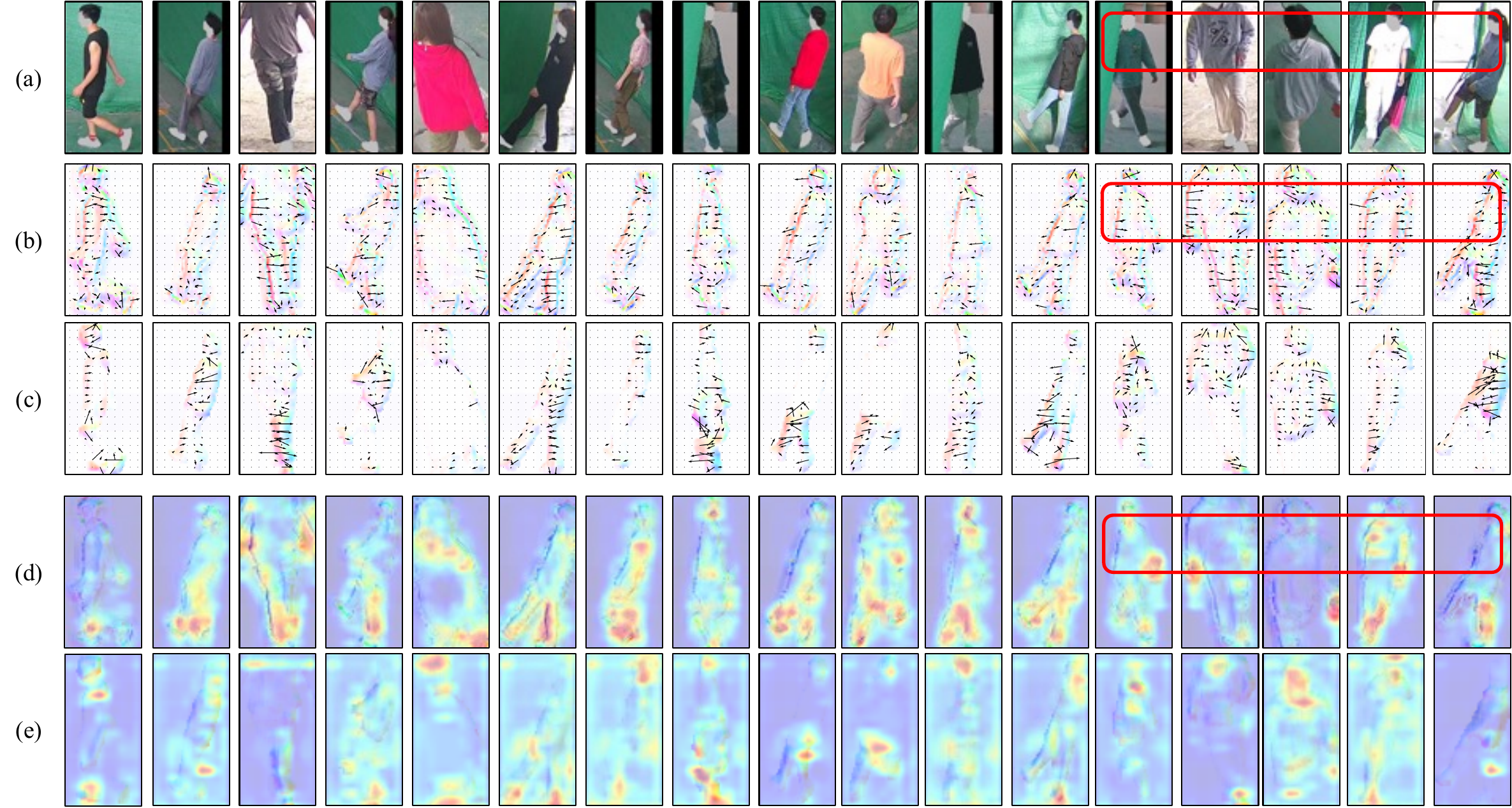}
\vspace{-0.5em}
\caption{
(a) Raw RGB images. 
(b) Static gait feature field, $G^{\text{Static}}$. 
(c) Dynamic gait feature field, $G^{\text{Dynamic}}$.
(d) Activation focus on $G^{\text{Static}}$. 
(e) Activation focus on $G^{\text{Dynamic}}$. 
(For optimal viewing, please refer to the color version and zoom in.)
}
\label{fig:vis}
\end{figure*}
\subsection{DenoisingGait} 
Building on the diffusion features $F_l$, the feature matching module aims to condense its multi-channel features at each pixel into a two-dimensional direction vector, thus reducing the expression of RGB-encoded noises while highlighting the local vectorized features of gait appearance and motion. 

As shown in Figure~\ref{fig:modules} (a), the within-frame and cross-frame matching follow a similar workflow: 
\begin{equation}
    \begin{aligned}
        F^Q_l &= S_l \cdot \mathcal{E}^Q(F_l), \\ 
        F^K_l &= S_{l + \Delta l} \cdot \mathcal{E}^K(F_{l + \Delta l}), \\ 
        G_l &= \mathcal{A} (F^Q_l, F^K_l)
    \end{aligned}
    \label{equ:matching}
\end{equation}
where $\Delta l$ denotes the temporal receptive scope and $\Delta l = 0$ for within-frame matching and $\Delta l > 0$ for cross-frame matching. 
The input $F_l \in \mathbb{R}^{\frac{H}{d} \times \frac{W}{d} \times 4}$ and $S_l \in \mathbb{R}^{\frac{H}{d} \times \frac{W}{d} \times 1}$ respectively represent the diffusion features and resized silhouette.
The modules $\mathcal{E}^{Q}$ and $\mathcal{E}^{K}$ are convolutional stacks, each containing four layers with identical architectures but independently trained weights (the input and output channel are respectively set to 4 and $C$). 
Here, $\mathcal{A}$ handles the assignment of directions (AoD) based on the query feature $F^Q_l \in \mathbb{R}^{\frac{H}{d} \times \frac{W}{d} \times C}$ and key feature $F^K_l\in \mathbb{R}^{\frac{H}{d} \times \frac{W}{d} \times C}$.
The resulting output $G_l \in \mathbb{R}^{\frac{H}{d} \times \frac{W}{d} \times 2}$ represent the gait feature field, where each pixel corresponds to a two-dimensional direction vector.

Next we focus on the AoD operation $\mathcal{A}$. 
For clarity, the following formulation ignores the subscript of $l$ (frame index) in Eq.~\ref{equ:matching}. 
For a single pixel $<i, j>$ within the query feature $F^Q$, represented as $f^Q_{<i, j>} \in \mathbb{R}^{1 \times C}$, we can identify its set of neighboring pixels within the key feature $F^K$ as shown in Figure~\ref{fig:modules} (b), : 
\begin{equation}
    \begin{aligned}
        \mathcal{M}^K_{<i, j>} =\{f^K_{<\hat{i}, \hat{j}>} & | \hat{i} \in \{i-\Delta h, ..., i + \Delta h \}, \\ & \hat{j} \in \{ j - \Delta w, ..., j + \Delta w \} \}, 
    \end{aligned}
    \label{equ:neighboring_pixels}
\end{equation}

For simplicity, we treat $\mathcal{M}^K_{<i, j>}$ as a matrix with elements arranged in raster order, \textit{i.e.}, from top to bottom and left to right.
Thus, $\mathcal{M}^K_{<i, j>}$ has a shape of $(2 \Delta h + 1) (2 \Delta w + 1) \times C$. 

Then, we compute the similarity distribution between $f^Q_{<i, j>}$ with its neighboring pixels: 
\begin{equation}
    \mathcal{P}_{<i, j>} = \text{SoftMax} (f^Q_{<i, j>} (\mathcal{M}^K_{<i, j>})^\top)
\end{equation}
where $\mathcal{P}_{<i, j>} \in \mathbb{R}^{1 \times (2 \Delta h + 1) (2 \Delta w + 1)}$ and the SoftMax function is conducted along the last dimension. 

To determine the final direction vector, we introduce a fixed direction template $\mathcal{T}$ as follows: 
\begin{equation}
    \begin{aligned}
        \mathcal{T} =\{[\hat{i}, \hat{j}] | &\hat{i} \in \{-\Delta h, ..., \Delta h \}, \\ & \hat{j} \in \{ - \Delta w, ..., \Delta w \} \}, 
    \end{aligned}
\end{equation}
Obviously, there is an element-wise correspondence between the direction template $\mathcal{T}$ and neighboring pixels $\mathcal{M}^K_{<i, j>}$ (in Eq.~\ref{equ:neighboring_pixels}). 
So we also treat $\mathcal{T}$ as a matrix with elements arranged in raster order, \textit{i.e.}, from top to bottom and left to right.
Thus, $\mathcal{T}$ has a shape of $(2 \Delta h + 1) (2 \Delta w + 1) \times 2$. 

Now, we assign pixel $<i, j>$ with a direction vector: 
\begin{equation}
    G_{<i, j>} = \mathcal{P}_{<i, j>} \mathcal{T}, 
\end{equation}

During each inference step, we perform both within-frame ($\Delta l = 0$) and cross-frame matching ($\Delta l > 0$) in parallel, resulting in both the static gait feature filed $G^{\text{Static}}$ and the dynamic gait feature field $G^{\text{Dynamic}}$. 

Interestingly, we observe that the magnitude of direction vector in the static gait feature field, $\| G^{\text{Static}}_{<i, j>} \|_2$, often reflects image texture intensity.
To enhance texture-invariant gait feature learning, we design a \textit{texture suppression} operation that applies random zero-padding to high-magnitude pixels during training with a probability $p$:
\begin{equation}
    G^{\text{Static}}_{<i, j>} = 
    \begin{cases}
        \textbf{0}, \qquad \text{if} \quad \| G^{\text{Static}}_{<i, j>} \|_2 > m \\
        G^{\text{Static}}_{<i, j>}, \qquad \text{otherwise}
    \end{cases}.
    \label{equ:texture}
\end{equation}
This operation encourages DenoisingGait to recognize that texture elements are unreliable, thereby promoting texture-free learning of gait features.

As illustrated in Figure~\ref{fig:pipeline} (c), the static and dynamic gait feature fields are fed in parallel into the subsequent GaitBase~\cite{fan2022opengait}. 
To fit its multi-branch inputs, we adopt the widely-used high-level attention fusion strategy proposed by Fan et al.~\cite{fan2023skeletongait}. 
Consistent with recent works~\cite{fan2020gaitpart, fan2023exploring, zheng2022gait3d, fan2022opengait, ma2023pedestrian, wang2023pointgait, wang2024cross, zou2024multi}, the training of DenoisingGait is driven by a combination of triplet loss and cross-entropy loss.

\begin{table*}[!t]
\centering
\vspace{-0.4em}
\caption{Within-domain Evaluation on CCPG~\cite{li2023depth} (CL: full cloth-changing, UP: up-changing, DN: pant-changing, and BG: bag-changing). }
 
\resizebox{1.7\columnwidth}{!}{ 
\setlength{\tabcolsep}{1mm}
\renewcommand{\arraystretch}{0.9}

\begin{tabular}{c|c|c|cccc|c|cccc|c} 
\toprule
\multirow{2}{*}{Input}    & \multirow{2}{*}{Model}                                     & \multirow{2}{*}{Venue} & \multicolumn{5}{c|}{Gait Evaluation Protocol}                                                               & \multicolumn{5}{c}{ReID Evaluation Protocol}                                                                 \\ 
\cmidrule{4-13}
                          &                                                            &                        & CL              & UP              & DN              & BG              & Mean            & CL              & UP              & DN              & BG              & Mean             \\ 
\cmidrule{1-13}
\multirow{4}{*}{Sils}     & GaitSet~\cite{Chao2019}                   & TPAMI'22               & 60.2           & 65.2           & 65.1           & 68.5           & 64.8            & 77.5           & 85.0           & 82.9           & 87.5           & 83.2             \\
                          & GaitPart~\cite{fan2020gaitpart}           & CVPR'20                & 64.3           & 67.8           & 68.6           & 71.7           & 68.1            & 79.2           & 85.3           & 86.5           & 88.0           & 84.8             \\
                          & GaitBase~\cite{fan2022opengait}           & CVPR'23                & 71.6           & 75.0           & 76.8           & 78.6           & 75.5            & 88.5           & 92.7           & 93.4           & 93.2           & 92.0             \\
                          & DeepGaitV2~\cite{fan2023exploring}        & Arxiv                  & 78.6           & 84.8           & 80.7           & 89.2           & 83.3            & 90.5       & \textbf{96.3}     & 91.4           & 96.7           & 93.7             \\ 
\cmidrule{1-13}
Flow                   & GaitBase$^f$                          & CVPR'23                & 70.0           & 74.5           & 77.7           & 77.5           & 74.9            & 82.4           & 88.9           & 90.9           & 91.5           & 88.4             \\ 
\cmidrule{1-13}
Sils + Parsing                  & XGait~\cite{zheng2024takes}                      & MM'24         & 72.8           & 77.0           & 79.1           & 80.5           & 77.4            & 88.3           & 91.8           & 92.9           & 94.3           & 91.9             \\ 
\cmidrule{1-13}
Sils + Parsing + Flow              & MultiGait++~\cite{jin2024exploring}                    & AAAI'25                & 83.9          & \textbf{89.0}           & 86.0           & 91.5           & 87.6            & -           & -           & -           & -           & -             \\ 
\cmidrule{1-13}
\multirow{2}{*}{Sils + Skeleton}& BiFusion~\cite{peng2024learning}      & MTA'23                    & 62.6           & 67.6           & 66.3           & 66.0           & 65.6            & 77.5           & 84.8           & 84.8           & 82.9           & 82.5             \\
                               & SkeletonGait++~\cite{fan2023skeletongait} & AAAI'24                & 79.1           & 83.9           & 81.7           & 89.9           & 83.7            & 90.2           & 95.0           & 92.9           & 96.9           & 93.8             \\ 
\cmidrule{1-13}
\multirow{1}{*}{RGB}      & BigGait~\cite{ye2024biggait} & CVPR'24 & 82.6           & 85.9           & 87.1           & 93.1           & 87.2            & 89.6           & 93.2           & 95.2           & 97.2           & 93.8             \\
\cmidrule{1-13}
\multirow{2}{*}{RGB+Sils} & GaitEdge~\cite{liang2022gaitedge}         & ECCV'22                & 66.9           & 74.0           & 70.6           & 77.1           & 72.2            & 73.0           & 83.5           & 82.0           & 87.8           & 81.6             \\ 
                          & DenoisingGait                             & Ours                   & \textbf{84.0}  & 88.0  & \textbf{90.1}  & \textbf{95.9}  & \textbf{89.5}   & \textbf{91.8}  & 95.8  & \textbf{96.4}  & \textbf{98.7}  & \textbf{95.7}  \\
                          
\bottomrule
\end{tabular}
}
\label{tab: ccpg_more}
\end{table*}

\subsection{Understanding Gait Feature Field}
\label{sec:visualization}
Figure~\ref{fig:vis} displays the static and dynamic gait feature fields, $G^{\text{Static}}$ and $G^{\text{Dynamic}}$, along with their activation focuses~\cite{selvaraju2017grad}. 

As observed, the static gait feature field $G^{\text{Static}}$ (Figure~\ref{fig:vis} (b)) reveals a gradient-like representation of gait appearance, with each pixel’s direction vector mostly oriented toward neighboring regions of high visual similarity: 
\begin{itemize}
    \item In related works~\cite{xu2022gmflow, shi2023videoflow, xu2023unifying}, similar local vectorized features are employed through supervised learning to capture local image depth or structural characteristics.
    Although the proposed DenoisingGait is driven solely by identity signals, we assume these local details may effectively populate $G^{\text{Static}}$ when they enhance human identification, guided by similar geometric constraints.
    \item As evidenced by the red boxes in Figure~\ref{fig:vis}, both the representation visualization and activation focus of $G^{\text{Static}}$ avoid texture-rich regions, thus encouraging texture-invariant gait feature learning.
\end{itemize}

As observed in Figure~\ref{fig:vis} (c) and (e), the dynamic gait feature field $G^{\text{Dynamic}}$ captures a flow-like representation of gait motion, with each pixel's direction vector primarily aligned with moving body parts. 
But unlike traditional optical flow, the dynamic gait feature field $G^{\text{Dynamic}}$ is fully recognition-oriented, with a focused objective to capture fine-grained gait motions for human identification. 
Additional video examples are provided in the \textbf{Supplementary Material}.

\section{Experiment}
\label{sec:exper}
\subsection{Dataset} 
In our experiments, we used three widely recognized clothing-changing and multi-view gait datasets: CCPG~\cite{li2023depth}, CASIA-B*\cite{yu2006framework}, and SUSTech1K\cite{Shen_2023_CVPR}, chosen for their provision of RGB images. Among these, CCPG~\cite{li2023depth} serves as the primary benchmark due to its extensive and diverse variations in clothing. This dataset includes a wide range of coats, pants, and bags in various colors and styles, with faces and shoes masked to simulate real-world challenges for cloth-changing gait recognition.

All implementations strictly adhere to the protocols set by dataset publishers. For performance reporting, we primarily follow the gait evaluation protocols for multi-view settings, using rank-1 accuracy as the main evaluation metric unless otherwise noted.

\subsection{Implementation Details}
\label{sec:implementation_details}
(1) All images are processed with Pad-and-Resize~\cite{ye2024biggait} to keep body proportion;
(2) We employ SD 1.5~\cite{rombach2022high} as the diffusion model for Figure~\ref{fig:pipeline};
(3) (H, W, d) = (768, 384, 8), (\(\Delta h\), \(\Delta w\)) = (3, 3), $m$=0.5 for Section~\ref{sec:method};
(4) The main hyper-parameters are listed in Table~\ref{tab:dataset};
(5) The SGD optimizer with an initial learning rate of 0.1 and weight decay of 0.0005 is utilized;
(6) During training, we adopt an ordered sampling strategy.

\begin{table}[t]
\centering
\caption{Implementation details. The batch size indicates the number of the IDs and the sequences per ID.}
\resizebox{0.9\columnwidth}{!}{
\setlength{\tabcolsep}{1mm} 
\begin{tabular}{ccccc}
\toprule
DataSet   & Batch Size & Schedule      & Frames       & Steps \\ \hline
CCPG      & (8, 4)     & (20k, 40k, 50k) & 20           & 60k         \\ 
CASIA-B*  & (8, 4)     & (15k, 25k, 35k) & 20           & 40k         \\
SUSTech1K & (8, 4)     & (15k, 25k, 35k) & 20           & 40k         \\
\bottomrule
\end{tabular}
}
\label{tab:dataset}
\end{table}

\begin{table*}[!t]
\centering
\vspace{-0.4em}
\caption{Within-domain Evaluation on CASIA-B*~\cite{yu2006framework, liang2022gaitedge} and SUSTech1K~\cite{Shen_2023_CVPR} (Abbreviations: NM for normal, BG for bag, CL for clothing, CR for carrying, UB for umbrella, UN for uniform, OC for occlusion, and NT for night).}
\resizebox{1.9\columnwidth}{!}{ 
\setlength{\tabcolsep}{1mm}
\renewcommand{\arraystretch}{0.9}
\begin{threeparttable}
\begin{tabular}{c|c|c|ccc|cccccccccc}
\toprule
\multirow{2}{*}{Input}         & \multirow{2}{*}{Model}& \multirow{2}{*}{Venue} & \multicolumn{3}{c|}{CASIA-B*}     & \multicolumn{10}{c}{SUSTech1K}                                                     \\ \cmidrule{4-16} 
                               &                       &                        & NM & BG & CL & NM & BG & CL & CR & UB & UN & OC & NT                   & R-1      & R-5                   \\\midrule
\multirow{4}{*}{Sils}          & GaitSet~\cite{Chao2019}                  &   TPAMI'22         & 92.3   & 86.1   & 73.4   & 69.1   & 68.2   & 37.4   & 65.0   & 63.1   & 61.0   & 67.2   & 23.0                     &   65.0       &   84.8            \\
                               & GaitPart~\cite{fan2020gaitpart}          &   CVPR'20          & 96.2   & 91.5   & 78.7   & 62.2   & 62.8   & 33.1   & 59.5   & 57.2   & 54.8   & 57.2   & 21.7                     &   59.2       &   80.8            \\
                               & GaitBase~\cite{fan2022opengait}          &   CVPR'23          & 96.5   & 91.5   & 78.0   & 81.5   & 77.5   & 49.6   & 75.8   & 75.5   & 76.7   & 81.4   & 25.9                     &   76.1       &   89.4            \\
                               & DeepGaitV2~\cite{fan2023exploring}       &   Arxiv            & 94.3   & 90.0   & 78.6   & 86.5   & 82.8   & 49.2   & 80.4   & 83.3   & 81.9   & 86.0   & 28.0                     &   80.9       &   91.9            \\\midrule
\multirow{2}{*}{Sils+Skeleton} & BiFusion~\cite{peng2024learning}         &   MTA'23           & 93.0   & 78.1   & 68.3   & 69.8   & 62.3   & 45.4   & 60.9   & 54.3   & 63.5   & 77.8   & 33.7                     &   62.1       &   83.4            \\
                               & SkeletonGait++~\cite{fan2023skeletongait}&   AAAI'24          & ---   & ---   & ---   & 85.1   & 82.9   & 46.6   & 81.9   & 80.8   & 82.5   & 86.2   & 47.5                     &   81.3       &   95.5            \\\midrule
Sils+Parsing+Flow              & MultiGait++~\cite{jin2024exploring}   &   AAAI'25          & --- & ---  & ---   & 92.0   & 89.4   & 50.4   & 87.6   & 89.7   & 89.1   & 93.4 & 45.1                &   87.4       &   95.6    \\ \midrule
RGB                            & BigGait~\cite{ye2024biggait}             &   CVPR'24          & \textbf{100.0}& 99.6  & 90.5   & 96.1   & \textbf{97.0}   & 73.2   & \textbf{97.2}   & 96.0   & 93.2   & \textbf{99.3}& \textbf{85.3}                &   \textbf{96.2}       &   \textbf{98.7}    \\ \midrule
RGB+Sils                       & DenoisingGait                         &   Ours             & \textbf{100.0}& \textbf{99.9}& \textbf{91.5}& \textbf{98.4}   & 96.3   & \textbf{79.0}   & 95.3  &  \textbf{97.1}  &  \textbf{94.7}  &  \textbf{99.3}  &   69.5  & 95.4         &  98.4             \\
\bottomrule
\end{tabular}
\end{threeparttable}
}
\label{tab:indomain}
\end{table*}

\subsection{Experimental Results}
\label{sec:main_results}
\noindent \textbf{Within-domain Evaluation. }
To show its superiority, DenoisingGait has been compared with various SoTA methods, including silhouette-based~\cite{Chao2019,fan2020gaitpart,gaitgl,fan2022opengait,fan2023exploring}, 
multimodal-based~\cite{peng2024learning,fan2023skeletongait,jin2024exploring,liang2022gaitedge}, and RGB-based methods~\cite{ye2024biggait}.

\begin{table}[!t]
\centering
\caption{
Cross-domain Evaluation, in which all methods are trained on CCPG and tested on CASIA-B* and SUSTech1K.
}
\vspace{-0.4em}
\renewcommand{\arraystretch}{1}

\resizebox{1.0\columnwidth}{!}{
\begin{tabular}{c|ccc|ccccc}
\multicolumn{9}{c}{\fontsize{12}{30}\selectfont Trained on \textbf{CCPG}} \\ 
\toprule
\multirow{3}{*}{Model} & \multicolumn{3}{c|}{CASIA-B*} & \multicolumn{5}{c}{SUSTech1K} \\ 
\cmidrule{2-9}
                       & NM    & BG    & CL    & NM    & BG    & CL    & UM    & Overall \\ 
\midrule
GaitSet~\cite{Chao2019}         & 47.4  & 40.9  & 25.8  & 11.5  & 14.5  & 8.2   & 11.0  & 12.8 \\
GaitBase~\cite{fan2022opengait} & 59.1  & 52.7  & 30.4  & 16.6  & 19.7  & 9.7   & 13.8  & 17.3 \\
BigGait~\cite{ye2024biggait}    & 77.4  & 71.5  & 33.6  & 60.7  & 57.2  & \textbf{43.7} & 41.1  & 56.4 \\
Ours                            & \textbf{83.9} & \textbf{76.1} & \textbf{34.8} & \textbf{66.9} & \textbf{59.7} & 37.3 & \textbf{45.7} & \textbf{59.1} \\ 
\bottomrule
\end{tabular}
}
\label{tab:cross}
\end{table}

As shown in Table~\ref{tab: ccpg_more}, DenoisingGait achieves superior performance across both gait recognition and ReID evaluation protocols on the CCPG~\cite{li2023depth} dataset. 
Under the gait recognition protocal, DenoisingGait shows improvements of +1.4\% on CL subset, +2.1\% on UP subset, +2.9\% on DN subset, +2.8\% on BG subset, and +2.3\% on average across all subsets. 
These results highlight DenoisingGait’s effectiveness in learning robust static and dynamic gait features by filtering out color and texture elements. 
The visualization of gait feature fields and their activation focus in Figure~\ref{fig:vis} also illustrate the robustness of DenoisingGait.

Table~\ref{tab:indomain} shows the within-domain evaluation on CASIA-B*~\cite{yu2006framework} and SUSTech1K~\cite{Shen_2023_CVPR}. 
In most cases, DenoisingGait outperforms other SoTA methods. 
However, its performance under night (NT) conditions on SUSTech1K~\cite{Shen_2023_CVPR} is somewhat limited, due to low-quality silhouettes in such settings. 
More details are provided in the Supplementary Material.
Even so, DenoisingGait achieves notable gains over other silhouette-based methods, demonstrating its robustness.

\begin{table}[!t]
\centering
\caption{
Ablation study on the knowledge-driven denoising: Diffusion (Ours) \textit{vs.} No Denoising (RGB Input) \textit{vs.} Color Denoising (Gray Image Input) \textit{vs.} DINOv2 and \textit{vs.} Sils+Flow.
}
\vspace{-0.4em}
\resizebox{0.9\columnwidth}{!}{ 
\begin{tabular}{c|c|ccccc}
\toprule
\multirow{2}{*}{Input}                                               & \multirow{2}{*}{\begin{tabular}[c]{@{}c@{}}Feature \\ Matching\end{tabular}} & \multicolumn{5}{c}{CCPG}                                                                     \\ \cmidrule{3-7}                                                                        
                                                                     &                           & CL                   & UP                   & DN                   & BG                   & R1 \\ \midrule
\multirow{2}{*}{\begin{tabular}[c]{@{}c@{}}Diffusion \\ Features\end{tabular}}& \checkmark          & \textbf{84.0}        & \textbf{88.0}        & \textbf{90.1}        & \textbf{95.9}        & \textbf{89.5}   \\
                                                                     & $\times$                  & 76.4                 & 79.8                 &  85.1                & 91.2                 &  83.1  \\ \midrule
\multirow{2}{*}{\begin{tabular}[c]{@{}c@{}}RGB\\ Image\end{tabular}} & \checkmark                & 77.9                 & 83.0                 &  87.7                & 94.4                 &  85.8  \\ 
                                                                     & $\times$                  & 62.2                 & 68.4                 &  75.7                & 82.5                 &  72.2  \\ \midrule
Gray Image                                                           & \checkmark                & 75.6                 & 81.4                 &  86.2                & 92.0                 &  83.8  \\ \midrule
\begin{tabular}[c]{@{}c@{}}DINOv2\\ Features\end{tabular}             & \checkmark                & 78.4                 & 83.2                 &  85.7                & 92.3                 &  84.9  \\ \midrule
Sils+Flow                                                            & N/A                       & 79.5                 & 83.1                 &  84.0                & 84.6                 &  82.8  \\
\bottomrule
\end{tabular}
}
\label{tab:knowledge-compare}
\end{table}

The CASIA-B*\cite{yu2006framework} and SUSTech1K\cite{Shen_2023_CVPR} datasets include limited clothing variations and do not mask faces or shoes. 
As noted by CCPG~\cite{li2023depth} dataset, evaluating RGB-based methods on these datasets may not accurately reflect real-world gait recognition performance, as algorithms can focus on the consistently visible features, such as face and shoes, which remain unchanged across samples.
We align with this viewpoint and therefore recommend the CCPG~\cite{li2023depth} dataset as the primary benchmark for evaluating RGB-based gait recognition methods.
Additionally, this work introduces cross-domain evaluation to further validate the effectiveness of DenoisingGait, as described below.

\noindent \textbf{Cross-domain Evaluation. }
As shown in Table~\ref{tab:cross}, cross-domain experiments (trained on CCPG, tested on CASIA-B* and SUSTech1K) further validate the generalization capability of DenoisingGait.  
We evaluated several SoTA methods based on silhouettes and RGB images. 
In most cases, DenoisingGait outperforms other SoTA methods. Specifically, it improves by +6.5\% in NM, +4.6\% in BG, and +1.2\% in CL scenarios on CASIA-B*, and achieves an overall improvement of +2.7\% on SUSTech1K. Although its CL performance has not reached the highest, it remains highly competitive.
More cross-domain experiments can be found in the \textbf{Supplementary Material}.

\begin{table}[!t]
\centering
\caption{Ablation study on \textit{feature matching} module.}
\vspace{-0.4em}
\resizebox{0.9\columnwidth}{!}{ 
\begin{tabular}{cc|ccccc}
\toprule
\multirow{2}{*}{\begin{tabular}[c]{@{}c@{}}Within-frame\\ Matching\end{tabular}} & \multirow{2}{*}{\begin{tabular}[c]{@{}c@{}}Cross-frame \\ Matching\end{tabular}} & \multicolumn{5}{c}{CCPG} \\ \cmidrule{3-7} 
                                                                                 &                                                                                  & CL   & UP   & DN   & BG   & R1 \\ \midrule
$\times$                                                                         & $\times$                                                                         & 76.4 & 79.8 & 85.1 & 91.2 & 83.1   \\
$\times$                                                                         & \checkmark                                                                       & 68.4 & 73.6 & 80.9 & 87.9 & 77.7   \\
\checkmark                                                                       & $\times$                                                                         & 82.0 & 86.5 & 89.9 & 95.4 & 88.5   \\
\checkmark                                                                       & \checkmark                         & \textbf{84.0} & \textbf{88.0} & \textbf{90.1} & \textbf{95.9} & \textbf{89.5}    \\
\bottomrule
\end{tabular}
}
\label{tab:matching-ablation}
\end{table}

\subsection{Ablation Study}
All ablation experiments are conducted on CCPG~\cite{li2023depth}. 

\noindent \textbf{Knowledge-driven Denoising by Diffusion Model. }
Table~\ref{tab:knowledge-compare} presents the results of incorporating DenoisingGait with various denoising knowledge: generative diffusion (Ours), no denoising (RGB input), color denoising (gray image input), DINOv2 representation~\cite{oquab2023dinov2}, and traditional gait representation like silhouette and optical flow. 
Compared to RGB images, Table~\ref{tab:knowledge-compare} demonstrates that the diffusion model effectively filters out gait-irrelevant cues in the videos. 
The performance improvement over grayscale images further emphasizes that the diffusion model not only removes color information but also filters out other non-essential cues.
While DINOv2 excels in representation learning, it is not specifically designed for denoising and thus show no improvements for DenoisingGait.
We also include two gait modalities, silhouette and optical flow. 
DenoisingGait outperforms GaitBase$^{s+f}$ with improvements of +4.5\% in the CL subset, +4.9\% in UP, +6.1\% in DN, +11.3\% in BG, and +6.7\% on average across all subsets. 
These results highlight the advantages of using the diffusion model as a knowledge-driven denoising module.

\noindent \textbf{Geometry-driven Denoising by feature matching. }
Table~\ref{tab:matching-ablation} shows that both within-frame and cross-frame feature matching contribute to the improvements of DenoisingGait. 
The visualization and discussion of the respective roles of within-frame and cross-frame matching for gait description can be found in Sec.~\ref{sec:visualization}. 
It is worth mentioning that using only cross-frame matching, \textit{i.e.}, taking only the dynamic gait feature fiedl $G^{\text{Dynamic}}$ as input (similar to GaitBase$^f$~\cite{fan2022opengait}, which uses the optical flow as input in Table~\ref{tab:indomain}), result in inferior performances. 
This aligns with the fact that static appearance features are crucial for gait description. 

Table~\ref{tab:within-ablation} presents additional ablation results on the within-frame feature matching mechanism. 
We conducted experiments to assess the impact of background removal (as defined in Eq.\ref{equ:matching}) and texture suppression (as described in Eq.\ref{equ:texture}). 
It is important to note that cross-frame feature matching was kept consistent throughout these experiments, and when background removal was excluded, cross-frame feature matching was also omitted.
Without the background removal operation, recognition performance drops noticeably, indicating that irrelevant background factors interfere with our geometry-driven denoising process, thereby damaging gait feature extraction. 
With the background removal operation, we found that applying the texture suppression operation significantly improves the Rank-1 accuracy across all scenarios, especially with a +3.0\% improvement in CL scenarios. 
This performance improvement indicates that the texture suppression operation effectively enhances DenoisingGait's texture-invariant gait feature learning.

\section{Challenges and Conclusions}
\label{sec:limitation}
\textbf{Challenges.}
Recent studies~\cite{ye2024biggait} highlight a pioneering trend of incorporating general knowledge of visual world into gait recognition. 
Related works often leverages large vision models, including discriminative models like DINOv2~\cite{oquab2023dinov2} and generative models such as the Latent Diffusion Models~\cite{rombach2022high} we use. 
However, these large models demand extensive training data, high computational costs, and substantial storage, all while advancing rapidly. 
Effectively utilizing and fairly comparing these diverse models, with due consideration of these costs, remains an unexplored issue for gait recognition.
Additionally, fully extracting useful features from walking videos while filtering out identity-irrelevant factors to create reliable gait pattern descriptions continues to be a persistent challenge for RGB-based gait methods~\cite{song2019gaitnet, zhang2020learning, ye2024biggait}.
In many cases, the proportion of cross-clothing pairs in the training data plays a critical role~\cite{ye2024biggait}, yet most of existing gait datasets contain only limited clothing changes.
\textbf{Supplementary Material} shows that DenoisingGait encounters a similar issue.

\noindent \textbf{Conclusions.}
The differences among pedestrians in videos are subtle since a video is in a very high dimensional space and with many noises. It makes gait recognition a very challenging task and is attracting increasing attentions. 
In our method, gait feature extraction is treated as a denoising process. This is achieved through the innovative use of diffusion model for knowledge-driven denoising and geometry-driven denoising by feature matching.
In addition to achieving a new state-of-the-art, the success of using a generative model for a challenging discriminative task highlights the potential of generative models. Furthermore, the geometry-driven denoising based on feature matching transforms features into gait feature fields, effectively masking gait-irrelevant cues while enhancing both structural characteristics and motion features.
We hope our work can inspire more researchers and encourage the exploration of additional gait denoising approaches in the future.
\begin{table}[!t]
\centering
\caption{More ablation results on \textit{Within-frame feature matching}.}
\vspace{-0.4em}
\resizebox{1.0\columnwidth}{!}{ 
\begin{tabular}{cc|ccccc}
\toprule
\multirow{2}{*}{\begin{tabular}[c]{@{}c@{}} Eq.~\ref{equ:texture}, Texture \\ Suppression  \end{tabular}} & \multicolumn{1}{c|}{\multirow{2}{*}{\begin{tabular}[c]{@{}c@{}} Eq.~\ref{equ:matching}, Background\\ Removal  \end{tabular}}} & \multicolumn{5}{c}{CCPG} \\ \cmidrule{3-7} 
                         &            & CL   & UP   & DN   & BG   & R1   \\ \midrule
$\times$   & $\times$    & 73.1 & 78.1 & 84.3 & 91.5 & 81.8 \\
\checkmark & $\times$   & 70.0 & 75.4 & 82.3 & 88.4 & 79.0   \\
$\times$   & \checkmark  & 81.0 & 86.5 & 89.7 & 95.7 & 88.2   \\
\checkmark    & \checkmark  & \textbf{84.0} & \textbf{88.0} & \textbf{90.1} & \textbf{95.9} & \textbf{89.5}    \\
\bottomrule
\end{tabular}
}
\label{tab:within-ablation}
\end{table}

\section*{Acknowledgements}
We thank Dingqiang Ye for his insightful discussions and help during the course of this work.
This work was supported by the National Natural Science Foundation of China under Grant 62476120, and the Shenzhen International Research Cooperation Project under Grant GJHZ20220913142611021. 


{
    \small
    \bibliographystyle{ieeenat_fullname}
    \bibliography{main}
}

\end{document}


\setcounter{figure}{4}
\twocolumn[{ %
\begin{center}
    {\bfseries \Large On Denoising Walking Videos for Gait Recognition}

    \vskip 1.5em
    \captionsetup{type=figure}
    \centering
    \includegraphics[width=1.8\columnwidth]{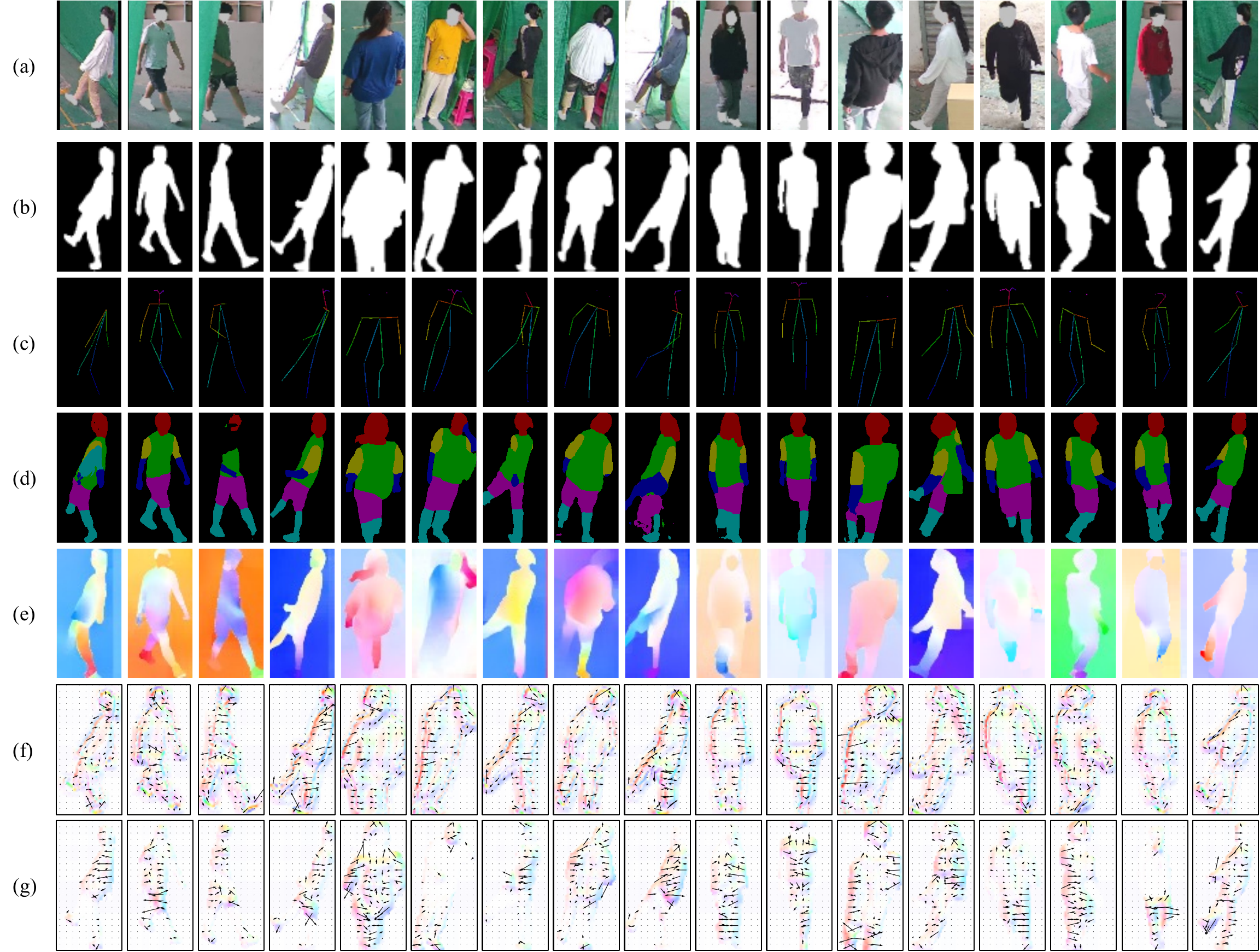}
    \caption{
    (a) Raw RGB images. 
    (b) Silhouette images. 
    (c) Skeleton images. 
    (d) Human parsing images. 
    (e) Optical flow images. 
    (f) Static gait feature field, $G^{\text{Static}}$. 
    (g) Dynamic gait feature field, $G^{\text{Dynamic}}$.
    (For optimal viewing, please refer to the color version and zoom in.)
    }
    \vspace{1em}
    \label{fig:sup_0}
\end{center}
}]

\begin{figure*}[t]
\centering
\includegraphics[width=1.7\columnwidth]{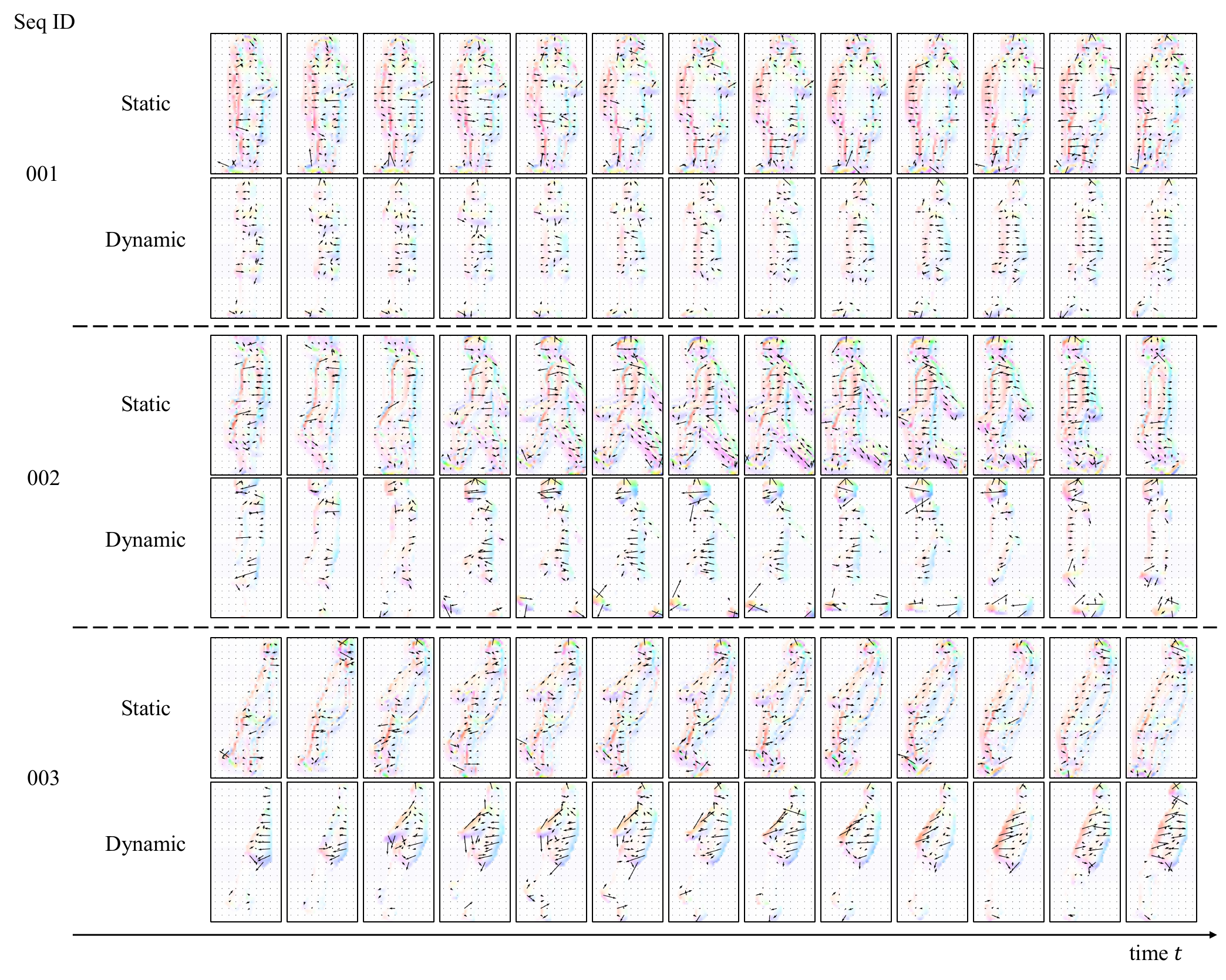}
\vspace{-0.9em}
\caption{The sequences of gait feature field, $G$}
\label{fig:supp_1}
\end{figure*}

\setcounter{table}{7}
\begin{table*}[!t]
\centering
\vspace{-0.4em}
\caption{Within-domain evaluation on CCPG~\cite{li2023depth} (CL: full cloth-changing, UP: up-changing, DN: pant-changing, and BG: bag-changing). }
 
\resizebox{1.8\columnwidth}{!}{ 
\setlength{\tabcolsep}{1mm}
\renewcommand{\arraystretch}{0.9}

\begin{tabular}{c|c|c|cccc|c|cccc|c} 
\toprule
\multirow{2}{*}{Input}    & \multirow{2}{*}{Model}                                     & \multirow{2}{*}{Venue} & \multicolumn{5}{c|}{Gait Evaluation Protocol}                                                               & \multicolumn{5}{c}{ReID Evaluation Protocol}                                                                 \\ 
\cmidrule{4-13}
                          &                                                            &                        & CL              & UP              & DN              & BG              & Mean            & CL              & UP              & DN              & BG              & Mean             \\ 
\cmidrule{1-13}
\multirow{3}{*}{Skeleton} & GaitGraph2~\cite{teepe2022towards}        & CVPRW'22               & 5.0            & 5.3            & 5.8            & 6.2            & 5.6             & 5.0            & 5.7            & 7.3            & 8.8            & 6.7              \\
                          & GaitTR~\cite{zhang2023spatial}           & ES'23                  & 15.7           & 18.3           & 18.5           & 17.5           & 17.5            & 24.3           & 28.7           & 31.1           & 28.1           & 28.1             \\
                          & SkeletonGait~\cite{fan2023skeletongait}              & AAAI'24                 & 29.0           & 34.5           & 37.1           & 33.3           & 33.5            & 43.1           & 52.9           & 57.4           & 49.9           & 50.8             \\ 
\cmidrule{1-13}
\multirow{5}{*}{Sils}     & GaitSet~\cite{Chao2019}                   & TPAMI'22               & 60.2           & 65.2           & 65.1           & 68.5           & 64.8            & 77.5           & 85.0           & 82.9           & 87.5           & 83.2             \\
                          & GaitPart~\cite{fan2020gaitpart}           & CVPR'20                & 64.3           & 67.8           & 68.6           & 71.7           & 68.1            & 79.2           & 85.3           & 86.5           & 88.0           & 84.8             \\
                          & AUG-OGBase~\cite{li2023depth}             & CVPR'23                & 52.1           & 57.3           & 60.1           & 63.3           & 58.2            & 70.2           & 76.9           & 80.4           & 83.4           & 77.7             \\
                          & GaitBase~\cite{fan2022opengait}           & CVPR'23                & 71.6           & 75.0           & 76.8           & 78.6           & 75.5            & 88.5           & 92.7           & 93.4           & 93.2           & 92.0             \\
                          & DeepGaitV2~\cite{fan2023exploring}        & Arxiv                  & 78.6           & 84.8           & 80.7           & 89.2           & 83.3            & 90.5       & \textbf{96.3}     & 91.4           & 96.7           & 93.7             \\ 
\cmidrule{1-13}
Flow                   & GaitBase$^f$                          & CVPR'23                & 70.0           & 74.5           & 77.7           & 77.5           & 74.9            & 82.4           & 88.9           & 90.9           & 91.5           & 88.4             \\ 
\cmidrule{1-13}
Sils + Parsing                  & XGait~\cite{zheng2024takes}                      & MM'24         & 72.8           & 77.0           & 79.1           & 80.5           & 77.4            & 88.3           & 91.8           & 92.9           & 94.3           & 91.9             \\ 
\cmidrule{1-13}
Sils + Flow              & GaitBase$^{s+f}$                    & CVPR'23                & 79.5           & 83.1           & 84.0           & 84.6           & 82.8            & 90.2           & 93.9           & 94.2           & 93.3           & 92.9             \\ 
\cmidrule{1-13}
\multirow{2}{*}{Sils + Skeleton}& BiFusion~\cite{peng2024learning}      & MTA'23                    & 62.6           & 67.6           & 66.3           & 66.0           & 65.6            & 77.5           & 84.8           & 84.8           & 82.9           & 82.5             \\
                               & SkeletonGait++~\cite{fan2023skeletongait} & AAAI'24                & 79.1           & 83.9           & 81.7           & 89.9           & 83.7            & 90.2           & 95.0           & 92.9           & 96.9           & 93.8             \\ 
\cmidrule{1-13}
\multirow{4}{*}{RGB}      & AP3D~\cite{gu2020appearance} & ECCV'20 & 53.4           & 57.3           & 69.7           & 91.4           & 67.8            & 62.6           & 67.6           & 82.0           & 97.3           & 77.4             \\
                          & PSTA~\cite{wang2021pyramid} & ICCV'21 & 42.2           & 52.2           & 60.3           & 84.5           & 59.8            & 51.9           & 62.0           & 72.3           & 94.1           & 70.1             \\
                          & PiT~\cite{zang2022multidirection} & TII'22 & 41.0           & 47.6           & 64.3           & 91.0           & 61.0            & 49.1           & 56.2           & 78.0           & 96.9           & 70.1             \\

                          & BigGait~\cite{ye2024biggait} & CVPR'24 & 82.6           & 85.9           & 87.1           & 93.1           & 87.2            & 89.6           & 93.2           & 95.2           & 97.2           & 93.8             \\
\cmidrule{1-13}
\multirow{2}{*}{RGB+Sils} & GaitEdge~\cite{liang2022gaitedge}         & ECCV'22                & 66.9           & 74.0           & 70.6           & 77.1           & 72.2            & 73.0           & 83.5           & 82.0           & 87.8           & 81.6             \\ 
                          & DenoisingGait                             & Ours                   & \textbf{84.0}  & \textbf{88.0}  & \textbf{90.1}  & \textbf{95.9}  & \textbf{89.5}   & \textbf{91.8}  & 95.8  & \textbf{96.4}  & \textbf{98.7}  & \textbf{95.7}  \\
                          
\bottomrule
\end{tabular}
}
\label{tab:ccpg_more}
\end{table*}

\begin{table*}[t]
\centering
\caption{
More cross-domain evaluation, where all methods are trained on one dataset and tested on the remaining two datasets.
}
\vspace{-0.5em}
\renewcommand{\arraystretch}{1.2}

\resizebox{1.3\columnwidth}{!}{
\begin{tabular}{c|cccc|ccccccc}
\multicolumn{12}{c}{\fontsize{12}{10}\selectfont (a) Trained on \textbf{CCPG}~\cite{li2023depth}} \\ 
\toprule[2pt]
\multirow{3}{*}{Model} & \multicolumn{9}{c}{Test Set}                                                                                   \\ \cmidrule{2-12}
                                & \multicolumn{4}{c|}{CASIA-B*} & \multicolumn{7}{c}{SUSTech1K} \\ \cmidrule{2-12}
                                & NM    & BG    & CL    & Overall & NM          & BG           & CL          &  UB           & UM           & OC          & Overall  \\
\midrule
GaitSet~\cite{Chao2019}         & 47.4  & 40.9  & 25.8  &  38.0  &  11.5       &  14.5        & 8.2         &  9.7         & 11.0         &  11.4       &  12.8           \\
GaitBase~\cite{fan2022opengait} & 59.1  & 52.7  & 30.4  &  47.4  &  16.6       &  19.7        & 9.7         &  11.8        & 13.8         &  16.8       &  17.3 \\
AP3D ~\cite{gu2020appearance}   & 53.7  & 46.2  & 11.9  &  37.3  &\textbf{68.1}&  52.4        & 36.2        &  42.6        &  38.3        &\textbf{65.9}&  55.3 \\
PSTA~\cite{wang2021pyramid}     & 49.7  & 42.3  & 8.8   &  33.6  &  51.4       &  37.8        & 25.7        &  33.8        &  26.8        &  52.5       &  40.6  \\
BigGait~\cite{ye2024biggait}    & 77.4  & 71.5  & 33.6  &  60.8  &  60.7       &  57.2        &\textbf{43.7}&  48.5        &         41.1 &  63.6       &  56.4 \\
Ours                            & \textbf{83.9} & \textbf{76.1} & \textbf{34.8}&\textbf{64.9} &66.9 &\textbf{59.7} &         37.3 &\textbf{55.0}& \textbf{45.7}& 64.0& \textbf{59.1}  \\ 
\bottomrule[2pt]
\end{tabular}
}

\vspace{0.5em}

\resizebox{2.0\columnwidth}{!}{

\fontsize{30}{35}\selectfont

\begin{tabular}{c|cccc|ccccc}
\multicolumn{10}{c}{\fontsize{35}{35}\selectfont (c) Trained on \textbf{SUSTech1K}~\cite{shen2023lidargait}}                   \\ 
\toprule[5pt]
\multirow{3}{*}{Model} & \multicolumn{9}{c}{Test Set}                                                                                   \\ \cmidrule{2-10}
                       & \multicolumn{4}{c|}{CASIA-B*}                                   & \multicolumn{5}{c}{CCPG}                     \\ \cmidrule{2-10}
                       & NM            & BG            & CL            & Overall       & CL            & UP            & DN            & BG            & Overall \\  \midrule
GaitSet                & 63.3          & 50.8          & 26.4          & 46.8          & 14.0          & \textbf{23.7} & 20.3          & 43.2          & 25.3    \\
GaitBase               & 73.1          & 61.2          & \textbf{28.2} & 54.2          & \textbf{16.8} & 21.7          & \textbf{26.0} & 42.7          & \textbf{26.8}           \\
AP3D                   & 56.7          & 48.1          & 15.3          & 40.0          & 5.5           & 7.9           & 13.9          & 35.1          & 15.6                    \\
PSTA                   & 31.2          & 27.7          & 10.6          & 23.2          & 3.7           & 5.7           & 9.5           & 26.5          & 11.4           \\
BigGait                & \textbf{91.1} & \textbf{85.8} & 18.7          & \textbf{65.2} & 4.5           & 11.5          & 11.9          & \textbf{45.5} & 18.4  \\
Ours                   & 87.0          & 81.6          & 21.1          & 63.2          & 5.5           & 11.0          & 15.4          & 45.3          & 19.3           \\

\bottomrule[5pt]
\end{tabular}

\quad

\begin{tabular}{c|ccccccc|ccccc}
\multicolumn{13}{c}{\fontsize{35}{35}\selectfont (b) Trained on \textbf{CASIA-B*}~\cite{yu2006framework}}                                                                                                              \\ 
\toprule[5pt]
\multirow{3}{*}{Model} & \multicolumn{12}{c}{Test Set}                                                                                                              \\  \cmidrule{2-13}
                       & \multicolumn{7}{c|}{SUSTech1K}                                 & \multicolumn{5}{c}{CCPG}                                                 \\   \cmidrule{2-13}
                       
               & NM            & BG            & CL            & UB           & UM          & OC          & Overall      & CL          & UP            & DN           & BG          & Overall        \\ \midrule
GaitSet        & 13.6          & 13.8          & 7.2           & 10.3         & 10.3        & 11.5        & 12.8         &\textbf{10.6}& 16.4          & 17.2         & 24.9        & 17.3          \\
GaitBase       & 19.2          & 16.7          & 8.1           & 12.0         & 14.5        & 15.6        & 15.6         &\textbf{10.6}& 18.1          & \textbf{21.4}& 28.7        & 19.7           \\
AP3D           & 60.3          & 44.2          & 29.3          & 42.6         & 49.5        & 56.3        & 48.3         & 2.1         & 2.9           & 3.9          & 6.1         &  3.8            \\
PSTA           & 47.4          & 33.2          & 19.9          & 25.5         & 33.0        & 43.4        & 34.6         & 1.7         & 1.9           & 3.4          & 5.0         &  3.0           \\
BigGait        & 68.6          & 62.8          &\textbf{36.9}  & 60.3         & 55.6        &\textbf{68.9}&\textbf{64.8} & 7.5         &\textbf{19.5}  & 14.2         &\textbf{43.0}&\textbf{24.6}  \\
Ours           &\textbf{69.8}  &\textbf{63.5}  &\textbf{36.9}  &\textbf{64.4} &\textbf{57.1}& 68.2        & 63.9         & 6.2         & 13.0          & 13.8         &  34.7       &  16.9            \\

\bottomrule[5pt]
\end{tabular}

}

\label{tab:cross_domain_all}
\vspace{-1em}

\end{table*}

\setcounter{section}{5}

\section{Supplementary Material}
\label{sec: Supplementary Material}
In this section, we first provide more details of Gait Feature Field. 
Then more experimental results under both the within and crossdomain scenarios are presented. 
Some related issues in rebuttal are attached as well.

\subsection{Understanding Gait Feature Field More}
\label{sec:understanding}
As illustrated in Figure~\ref{fig:sup_0}, there are various vision modalities commonly used for gait description, including (but not limited to) binary silhouettes, skeleton coordinates, human parsing, and optical flow images.
Typically derived from RGB videos, these modalities are provided by third-party tasks designed to express specific physical meanings, such as separating background from body regions or capturing joint-level and pixel-level walking movements. 
While these modalities effectively exclude gait-unrelated cues, it is important to note that their definitions are not explicitly tailored for identifying individuals based on gait.
Many end-to-end works~\cite{song2019gaitnet, zhang2020learning, ye2024biggait} argued this point and highlighted the superiority of global optimization in directly extracting gait characteristics from pedestrian videos. 

In this study, the comparison between traditional optical flow and our dynamic gait feature field, both designed to capture pixel-level temporal dynamics, 
highlights a significant distinction between gait representations generated by third-party tasks and those specifically optimized for gait recognition.
As illustrated in Figure~\ref{fig:sup_0} (e), optical flow images effectively depict smooth dynamics across the entire body. 
In contrast, the proposed dynamic gait feature field, illustrated in Figure~\ref{fig:sup_0} (g), adaptively adjusts the scale of movements at the pixel level.
Because its learning process is entirely driven by recognition supervision, we hypothesize that the dynamic gait feature field effectively amplifies identity-related movements while suppressing those unrelated to identity.

Similarly, our static gait feature field, shown in Figure~\ref{fig:sup_0} (f), captures the vectorized local details of gait appearance, with high-magnitude pixels predominantly concentrated along the body’s edge regions.
This phenomenon aligns with the characteristics of human silhouettes and parsing images, as these edge regions effectively convey body and part shape features essential for gait understanding. 
Moreover, the vector-valued nature of the static gait feature field makes it more informative than traditional appearance-based gait modalities.

In summary, DeonisingGait effectively extracts recognition-oriented features by leveraging the proposed knowledge- and geometry-driven gait denoising priors.

\subsection{More Experimental Results}
\noindent \textbf{More Within-domain Evaluation on CCPG.}
In Table~\ref{tab:ccpg_more}, in addition to the content in the main text, we include several video-based ReID methods, including AP3D~\cite{gu2020appearance}, PSTA~\cite{wang2021pyramid}, and PiT~\cite{zang2022multidirection}.
Compared to these methods, DenoisingGait outperforms them considerably, e.g., +30.6\% for cloth-changing (CL), +30.7\% for up-changing (UP), +20.4\% for pant-changing (DN), and +4.5\% for bag-changing (BG) scenarios.
Accordingly, we consider that DenoisingGait can effectively remove gait-irrelevant cues from RGB videos and extract robust identity representations.

\noindent \textbf{More Cross-domain Evaluation. }
In addition to comparing DenoisingGait with several state-of-the-art (SoTA) gait recognition methods, we include two video-based ReID methods, AP3D~\cite{gu2020appearance} and PSTA~\cite{wang2021pyramid}, as references.
Table~\ref{tab:cross_domain_all} presents cross-domain experiments conducted on CCPG, CASIA-B*, and SUSTech1K, where the model is trained on a certain dataset and evaluated on the other two.

The results reveal phenomena similar to those reported in previous studies~\cite{ye2024biggait}. Specifically, Table~\ref{tab:cross_domain_all} illustrates how DenoisingGait’s cross-domain performance varies depending on the training dataset. When trained on CCPG, DenoisingGait demonstrates strong adaptability to unseen datasets, outperforming both video-based ReID methods~\cite{gu2020appearance, wang2021pyramid}, silhouette-based methods~\cite{Chao2019, fan2022opengait}, and the RGB-based method~\cite{ye2024biggait}.

However, when trained on CASIA-B* or SUSTech1K, DenoisingGait encounters challenges in cross-dressing scenarios on CCPG, particularly in settings such as CL, UP, and DN. Table~\ref{tab:cross_domain_all} presents the cross-domain experiments conducted on CCPG, CASIA-B*, and SUSTech1K, where the model is trained on one dataset and tested on the other two datasets.

This limitation, fortunately, can be addressed with more diverse training data. 
Compared to CASIA-B* and SUSTech1K, CCPG offers a broader range of outfit variations. 
As shown in Table~\ref{tab:cross_domain_all} (a), training on CCPG allows DenoisingGait to develop more robust gait representations. 
In summary, the distribution of training data influences learned representations. 
Greater cross-dressing diversity improves performance in such scenarios, though achieving strong cross-dressing capability with limited diversity remains an open challenge.

\subsection{Related Issues in Rebuttal}
\noindent \textbf{Q1: Comparison to Multi-Inputs.}
We developed multi-input (RGB+Sil) GaitBase~\cite{fan2022opengait} and BigGait~\cite{ye2024biggait}, where GaitBase uses silhouette-masked RGB and BigGait replaces the learned mask with silhouettes.
Apart from this, the settings remain consistent with the original GaitBase and BigGait.
As shown in Table~\ref{tab:multi-input}, DenoisingGait remains the best performance on CCPG, while RGB+Sil BigGait performs even worse. 
We suspect that this drop may be due to the strong shape priors within silhouettes, which could prevent BigGait from learning better features from DINOv2.
\begin{table}[h!]
\centering
\caption{Comparison to multi-inputs on CCPG}
\vspace{-0.5em}
\resizebox{1.0\columnwidth}{!}{%
\begin{tabular}{c|c|cccc|c}
\hline
CCPG        & Input type & CL            & UP            & DN            & BG            & R1            \\ \hline
GaitBase~\cite{fan2022opengait}     & RGB+Sil    & 74.4          & 80.1          & 87.1          & 93.2          & 83.7          \\
BigGait~\cite{ye2024biggait}      & RGB        & 82.6          & 85.9          & 87.1          & 93.1          & 87.2          \\
BigGait~\cite{ye2024biggait}      & RGB+Sil    & 78.0          & 82.0          & 86.5          & 92.8          & 84.8          \\
GaitEdge~\cite{liang2022gaitedge}     & RGB+Sil    & 66.9          & 74.0          & 70.6          & 77.1          & 72.2          \\
DenoisingGait & RGB+Sil    & \textbf{84.0} & \textbf{88.0} & \textbf{90.1} & \textbf{95.9} & \textbf{89.5} \\ \hline
\end{tabular}
}
\label{tab:multi-input}
\end{table}

\noindent \textbf{Q2: Justification for timestep \textit{t}=700 in knowledge-driven denoising}. 
Much evidence suggests that early timesteps (\textit{t}→\textit{T}) in diffusion models mainly capture overall shapes, while later timesteps (\textit{t}→0) focus on refining details~\cite{ke2024repurposing,ren2024tiger,chen2024tino,tian2024diffuse}. 
Based on this, we set timestep \textit{t}=700 to retain overall shape features and partially mitigate identity-unrelated RGB details, as validated in Figure 2 (b).
As shown in Table~\ref{tab:timestepsustech1k}, more experiments on SUSTech1K confirm the effectiveness of timestep \textit{t}=700, consistent with observations from CCPG in Figure 2 (b). 
Here, we focus on the challenging cloth-changing (CL) cases on both CCPG (Figure 2 (b)) and SUSTech1K (Table~\ref{tab:timestepsustech1k}), where the color and texture of cloth become significant noise for gait recognition.

\begin{table}[h!]
\centering
\caption{Comparing Rank-1 Accuracy of different timestep \textit{t} in Baseline on SUSTech1K.}
\vspace{-0.5em}
\resizebox{0.9\columnwidth}{!}{%
\begin{tabular}{c|c|c|c|c|c}
\hline
SUSTech1K & \textit{t}=1000 & \textit{t}=700         & \textit{t}=500         & \textit{t}=300 & \textit{t}=100 \\ \hline
NM-cases        & 97.5   & 97.4 & \textbf{97.6}           & 97.2  & 96.7  \\ 
CL-cases        & 68.6   & \textbf{76.5}          & 75.4   & 74.8  & 69.1  \\
Mean-R1         & 94.6   & \textbf{95.1} & \textbf{95.1}   & 94.4  & 93.5  \\
\hline
\end{tabular}
}
\label{tab:timestepsustech1k}
\end{table}

\noindent \textbf{Q3: About multi-timestep input}.
For the multi-timestep input, we test \textit{t}=\{700, 500\} and \textit{t}=\{700, 500, 300\} on CCPG. The Mean-R1 Accuracy improved by +0.6\% and +0.9\%, respectively, while the time costs increased to 2$\times$ and 3$\times$. 

\noindent \textbf{Q4: NT (night)-cases on SUSTech1K}.
The NT-cases silhouette quality of SUSTech1K is poor.
Table~\ref{tab:sustech1k-sil} presents experiments conducted on SUSTech1K.
In this case, DenoisingGait outperforms GaitBase by +43.6\%, showing its robustness under low-visibility conditions.
Once we enhance the SUSTech1K silhouette quality (denoted by *, in Table~\ref{tab:sustech1k-sil}), DenoisingGait improves from 69.5\% to 90.2\%, surpassing BigGait's 85.3\%. 
Meanwhile, GaitBase improves from 25.9\% to 68.9\%. 
\begin{table}[h!]
\centering
\caption{Comparing Rank-1 Accuracy on SUSTech1K.}
\vspace{-0.5em}
\resizebox{1.0\columnwidth}{!}{%
\begin{tabular}{c|c|c|c|c|c}
\hline
SUSTech1K & GaitBase & GaitBase* & Ours & Ours* & BigGait \\ \hline
NT-cases  & 25.9     & 68.9          & 69.5      & \textbf{90.2}  & 85.3    \\
Mean-R1   & 76.1     & 85.2          & 95.4      & \textbf{97.5}  & 96.2    \\ \hline
\end{tabular}
}
\label{tab:sustech1k-sil}
\end{table}

\noindent \textbf{Q5: The latent space $F_l$ and Gait Feature Field can be noisy}. 
Traditional gait inputs can also be noisy, \textit{e.g.}, silhouette and parsing images often retains clothing shapes and even background, especially in in-the-wild imagery. 
DenoisingGait is designed to progressively remove identity-unrelated cues. 
While $F_l$ only partially mitigate RGB noises, Feature Matching is further introduced to enhance denoising.
We believe DenoisingGait's advantage is not from texture or color, as evidenced by: 

\noindent a) In Table~\ref{tab:ccpg_more}, it outperforms BigGait, while the BigGait~\cite{ye2023gait} significantly surpasses ReID methods, despite the latter having full access to color and texture cues. 

\noindent b) In Table 5, with denoising via Diffusion Features and Feature Matching, DenoisingGait achieves the best performance. 

\noindent c) Feature and activation visualizations (Figure 4) further support this conclusion.

\noindent \textbf{Q6: Vectors Pointing out of Body}.
These vectors are mainly located within the dynamic $G^{\text{Dynamic}}$ (shown in Figure 4), revealing body movements (videos are in Figure~\ref{fig:supp_1}).
The directions are totally determined by neighboring visual similarity. 
Section 3.3 and \ref{sec:understanding} can provide more understandings. 

\noindent \textbf{Q7:About Global Matching}. 
Integrating global matching into DenoisingGait yielded a slight 0.3\% improvement.
We assume that Local Matching, widely used in related works~\cite{lowe2004distinctive,xu2023unifying}, allows the CNN head to capture both local and global cues.

\noindent \textbf{Q8: Visualize BG cases}. 
As shown in Figture~\ref{fig:bg_case}, DenoisingGait is still robust in this case (activations are not on BG regions, below). 
\begin{figure}[!h]
  \centering
   \includegraphics[width=1.0\linewidth]{rebuttal/bg_case.pdf}
    \vspace{-2em}
    \caption{
    (a) Raw RGB image.
    (b) Static gait feature field, $G^{\text{Static}}$. 
    (c) Activation focus on $G^{\text{Static}}$. 
    }
    \label{fig:bg_case}
\end{figure}



{
    \small
    \bibliographystyle{ieeenat_fullname}
    \bibliography{main}
}